\documentclass[11pt]{article}
\usepackage[T1]{fontenc}
\usepackage[utf8]{inputenc}
\usepackage[margin=1in]{geometry}
\usepackage{graphicx}
\usepackage{booktabs}
\usepackage{array}
\usepackage{ragged2e}
\usepackage[htt]{hyphenat}
\newcolumntype{L}[1]{>{\RaggedRight\arraybackslash}p{#1}}
\usepackage{caption}
\usepackage[section]{placeins}
\usepackage[colorlinks=true,allcolors=blue]{hyperref}
\title{The MODA General Attribute Suite:\\A Four-Track Evaluation Benchmark for Fashion Attribute Extraction}
\author{Arkid Mitra\\Hopit AI\\\texttt{https://github.com/hopit-ai/Moda\_ner}}
\date{}

\begin{document}
\maketitle

\begin{abstract}
Fashion attribute extraction is evaluated inconsistently: results are reported as single aggregate numbers across image types that pose different problems, fields that are not visible in an image are scored as ordinary negatives, and the effect of vocabulary mismatch between datasets is acknowledged but not measured. We release the MODA General Attribute Suite, a four-track benchmark that keeps these problems separate by construction. Each track (localized garment crops, catalogue product images, applicability-aware full-body photographs, and product text) carries its own frozen test set, input contract, metric, and leakage unit, and the tracks are never averaged. The protocol requires label-blind prediction, SHA-256 commitment of prediction files before any label is opened, a fail-closed scorer, and a 10,000-sample paired cluster bootstrap at each track's natural leakage unit; promotion requires a positive interval on every track rather than a favourable mean. We release the scorers, the split builders, our own prediction files with their hashes for every image-track system in the main results and for runs we lose, and the MODA\_NER(V) model checkpoints for three of the four tracks. The text model is not distributed; its benchmark is. We report baseline results, six interventions that did not improve them, and the limitations of the evidence. On the garment-crop track we also measure how far the headline depends on the benchmark's own design: the order of the two strongest systems depends on whether fields are micro- or macro-averaged, two thirds of the cells the protocol judges carry no annotation, and disabling the applicability decision cuts attribute micro-F1 from 0.63 to 0.27, so most of that score rests on absence decisions the source labels cannot adjudicate. This version also corrects one figure in an external comparison that an inference preprocessing defect had affected.
\end{abstract}
\section{Introduction}

Turning a fashion photograph into structured product data (category, silhouette, sleeve length, neckline, fabric, pattern) is a routine industrial requirement and an unusually awkward evaluation problem. Three difficulties recur.

\textbf{Input contracts differ, and averaging hides it.} A cropped garment on white, a studio catalogue shot, and a full-body street photograph are different tasks. A model can be strong on one and weak on another, and a single headline number lets the strong result absorb the weak one.

\textbf{Not every attribute is present in every image.} A full-body photograph may show no outer garment at all. Scoring \texttt{outer\_fabric} as an ordinary misclassification conflates two different failures: not knowing the fabric, and not knowing there is no coat. The second is the one that produces confident nonsense in a production catalogue.

\textbf{Vocabularies disagree.} Two corpora that both annotate ``neckline'' partition it differently. Mapping between them is lossy in ways that are widely acknowledged and rarely quantified.

Underneath all three is one claim: different fashion datasets do not simply constitute more training examples of the same attribute extraction task. They encode different contracts about what is visible, applicable, localized, and labelable. A benchmark that pools them measures the average of several different questions. The suite addresses the first two difficulties by construction and measures the third.

\section{Related work}

\textbf{Fashion attribute datasets.} DeepFashion [10] and the million-scale iMaterialist Fashion Attribute Dataset [9] established large-scale attribute annotation for clothing, and the three corpora this suite builds on (Fashionpedia [1], DeepFashion-MultiModal [2] and Shopping100k [3]) each define their own vocabulary, their own image contract, and their own view of which attributes are annotated at all. That heterogeneity is usually treated as an obstacle to pooling. We treat it as the object of measurement.

\textbf{Fashion vision-language models.} FashionCLIP [4] adapted CLIP-style contrastive pretraining to the domain, and Marqo-FashionSigLIP [5] fine-tuned a SigLIP backbone [6] for fashion retrieval; general-purpose VLMs such as Qwen3-VL [7] are increasingly applied zero-shot. These are the systems we evaluate against rather than extend.

\textbf{Evaluation discipline.} Kapoor and Narayanan [13] document leakage as a systemic cause of irreproducible ML results across disciplines, which motivates our split at each track's natural leakage unit and our hash-commitment of predictions before labels are opened. LookBench [15] takes contamination seriously in the adjacent fashion setting of image retrieval, refreshing its evaluation set over time; our tracks are frozen instead, and reproducibility rests on published prediction files rather than on novelty of the images.

\textbf{Label-space mismatch.} The closest prior work is Label-Aligned Transfer [12], which projects annotations from heterogeneous source datasets into a target label space for object detection. Cross-domain attribute recognition has a longer history in fashion [11]. Both treat vocabulary mismatch as an obstacle to be \emph{repaired} by a method. Section 8 instead \emph{measures} what it costs when left unrepaired, on a frozen benchmark, and shows it can invert the ranking of two systems.

\textbf{Applicability.} We adopt the three-tier framework of Shukla and Sonalkar [8]. Tier-2 is related to selective classification [14], where a model abstains on inputs it is unsure of, but the question differs: tier-2 asks whether the attribute is present in the image at all, not whether the model is confident about it. Absence is a property of the garment, not of the classifier.

\section{The suite}

\begin{center}\small
\begin{tabular}{lL{0.242\linewidth}L{0.174\linewidth}ll}
\toprule
Track & Frozen test & Input contract & Fields & Leakage unit \\
\midrule
\texttt{crop} & 4,688 garment crops / 1,158 images [1] & oracle garment crop & 15 & source image \\
\texttt{catalog} & 9,995 images / 61,384 cells [3] & catalogue product image & 10 & image \\
\texttt{fullbody} & 5,000 images / 1,751 product groups [2] & full-body photograph & 18 + explicit N/A & product group \\
\texttt{text} & 1,071 rows (765 standard, 306 hard) & product title or description & 13 entity types & row \\
\bottomrule
\end{tabular}
\end{center}

Three of the four tracks ship a released checkpoint. The \texttt{text} track ships its builder, scorer and our prediction file, but not its weights.

Tracks are selected by input contract, not by score. \texttt{crop} covers the category hierarchy, silhouette, sleeve, neckline, collar, closure, hemline, waist, pattern, material and surface treatment, but carries no colour or fit and is supplied with localization. \texttt{catalog} adds colour and fit but has no applicability or accessory fields. \texttt{fullbody} carries region-specific fabric and pattern, accessories, and explicit not-applicable states, over a closed vocabulary. \texttt{text} scores exact character spans.

Because the tracks are not comparable, the suite refuses to average them, and the promotion gate is conjunctive: a system is promoted only if the 95\% interval lower bound is above zero on every track it claims.

\subsection{Three-tier scoring on \texttt{fullbody}}

\texttt{fullbody} reports three numbers rather than one:

\begin{itemize}
\item \textbf{tier-1}, macro-F1 over all 18 fields including the not-applicable class;
\item \textbf{tier-2}, F1 on the applicability decision alone, that is, whether the system knows the attribute is not visible;
\item \textbf{tier-3}, macro-F1 restricted to rows where the attribute is genuinely present.
\end{itemize}

This decomposition is not ours. We adopt the three-tier framework of Shukla and Sonalkar [8], who introduced it to evaluate zero-shot vision-language models on DeepFashion-MultiModal, and apply it to supervised routes. Tier-2 is the metric most attribute work still omits: a system can be respectable on tier-1 while being unable to tell absence from ignorance, and that difference decides whether its output is safe to write into a catalogue unattended.

\section{Protocol}

The protocol is a fixed pipeline, not a checklist: each stage gates the next, and two stages (freezing the manifest, and committing predictions before labels are opened) are hard gates that cannot be skipped or reordered. Figure 1 shows the sequence; the paragraph below states what each stage guarantees.

\begin{figure}[!tbp]
\centering
\includegraphics[width=\linewidth]{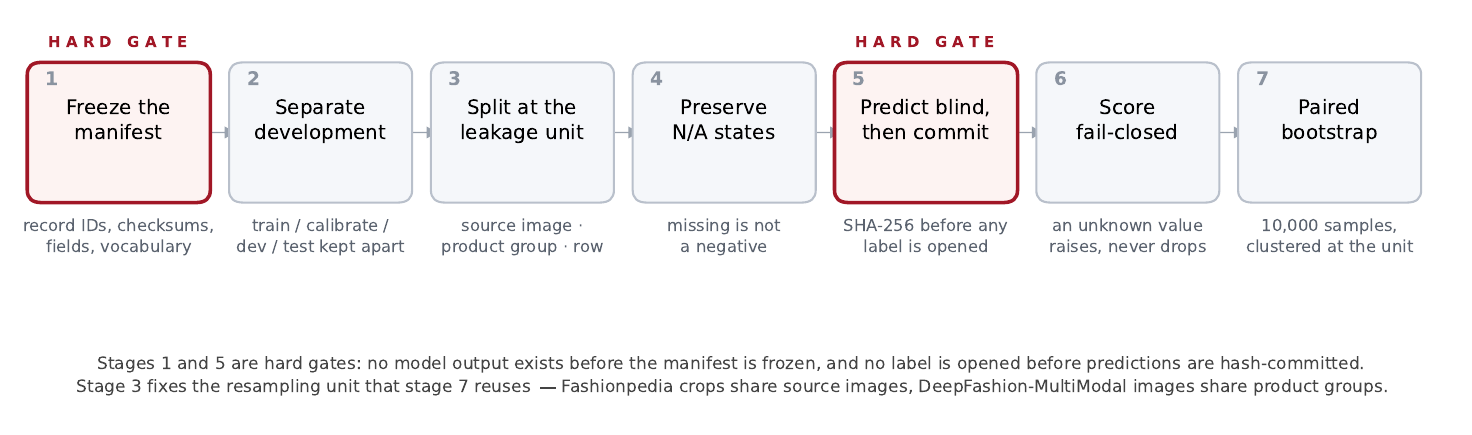}
\caption{The evaluation protocol as a seven-stage gated pipeline. Stages 1 and 5 are hard gates: no model output exists before the manifest is frozen, and no label is opened before predictions are hash-committed. Fashionpedia [1] crops share source images; DeepFashion-MultiModal [2] images share product groups, splitting on rows instead would leak, which is why stage 3 fixes the unit that stage 7's bootstrap later reuses.}
\end{figure}

We do not redistribute source corpora or their labels. Each track ships a builder that reconstructs the frozen split from record IDs and checksums against the corpus obtained from its original source under that source's terms.

\section{Baseline results}

\begin{center}\small
\begin{tabular}{L{0.198\linewidth}rrr}
\toprule
Track / metric & MODA & Comparator & Paired 95\% CI \\
\midrule
\texttt{crop}, attribute micro-F1 & \textbf{0.6300} & FashionSigLIP [5] 0.6245 & [+0.0014, +0.0097] \\
\texttt{crop}, attribute field-macro F1 & 0.6074 & FashionSigLIP [5] 0.6093 & [$-$0.0077, +0.0040] \\
\texttt{catalog}, field-macro set F1 & \textbf{0.8292} & FashionCLIP 2.0 [4] 0.6657 & [+0.1595, +0.1676] \\
\texttt{fullbody} tier-1 & \textbf{0.6917} & FashionCLIP 2.0 0.5943 & [+0.0891, +0.1053] \\
\texttt{fullbody} tier-2 (N/A) & \textbf{0.6637} & FashionCLIP 2.0 0.6088 & [+0.0433, +0.0657] \\
\texttt{fullbody} tier-3 (visible) & \textbf{0.5785} & FashionCLIP 2.0 0.4969 & [+0.0723, +0.0905] \\
\texttt{text}, strict span F1 & \textbf{0.8723} & n/a & n/a \\
\bottomrule
\end{tabular}
\end{center}

Attribute micro-F1 is the \texttt{crop} track's pre-registered headline. Field-macro F1 is reported beside it because the two disagree: on field-macro the two systems cannot be separated (Section 6.1).

The \texttt{text} row is not comparable to the others: no external comparator was run, its labels are cleaned silver rather than human gold, and its model is not released.

Field by field, the margin is not carried by a handful of attributes: \texttt{catalog} wins all 10 fields, mean relative gain +30.8\%, widest on \texttt{fabric} (0.6693 against 0.3527) and \texttt{neckline} (0.8337 against 0.5228); \texttt{fullbody} wins 17 of 18, mean relative gain +18.7\%, the single loss being \texttt{lower\_fabric} at 0.3287 against 0.3329, where both systems perform poorly. Figure 2 shows every field on both tracks.

\begin{figure}[!tbp]
\centering
\includegraphics[width=\linewidth]{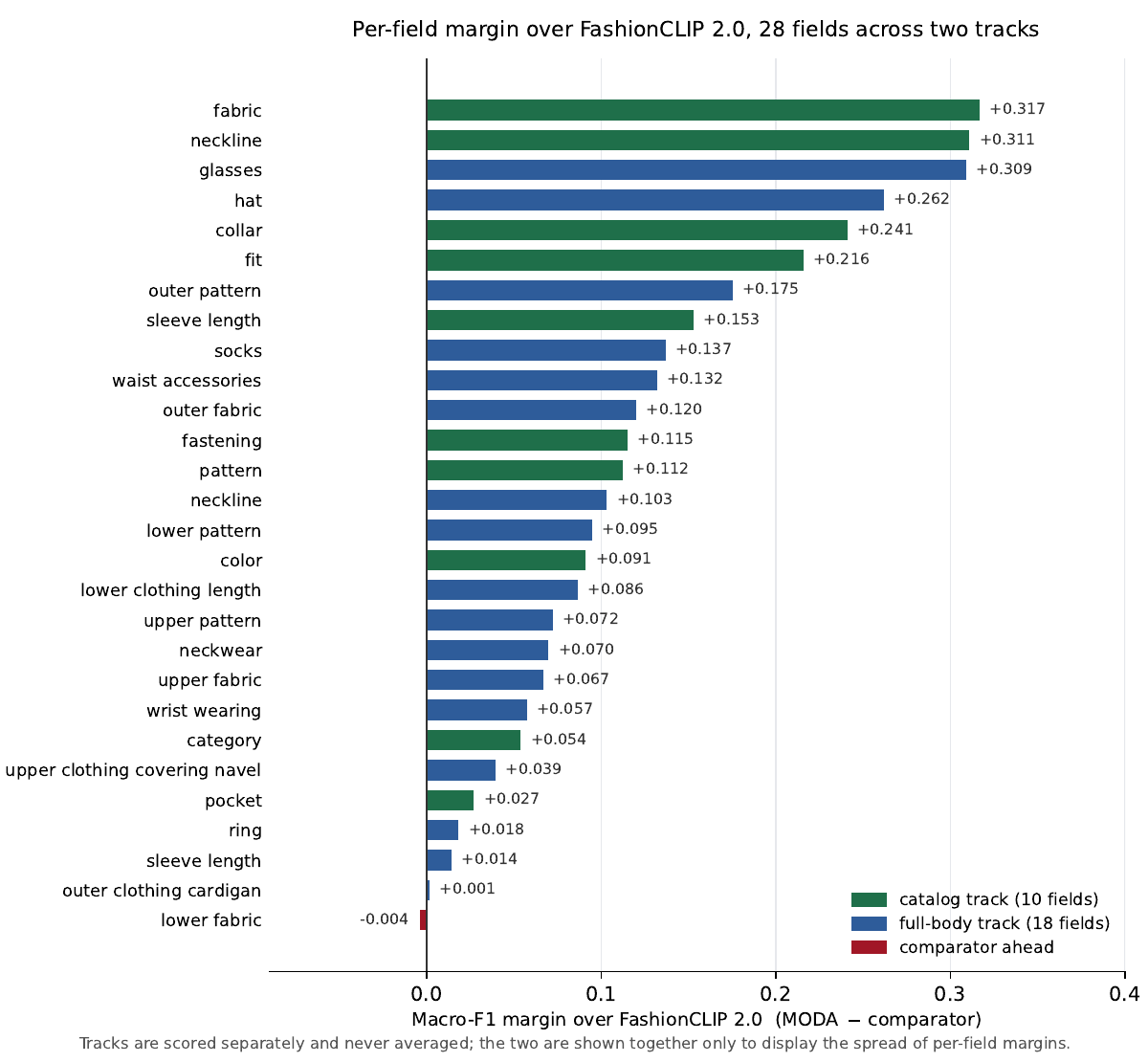}
\caption{The margin is distributed, not carried by outliers: 27 of 28 fields are positive, and the median margin is around +0.10. The two tracks are scored separately and never averaged; they appear on one axis here only to show the spread. The single negative field, \texttt{lower\_fabric}, is one where neither system is usable.}
\end{figure}

The \texttt{crop} row requires a caveat we state rather than bury. Its 0.6245 comparator uses our own architecture and training release with a frozen third-party encoder substituted in: it is an encoder ablation, not a comparison against another organisation's product. Genuine third-party zero-shot \texttt{crop} baselines [5,7] score 0.1805 and 0.1817, and neither was built for this task.

Weak fields, stated plainly: \texttt{material} 0.4148 on \texttt{crop}; \texttt{fabric} 0.6693 and \texttt{fit} 0.6708 on \texttt{catalog}; collar style, neckline and rare applicability values throughout. Using F1 >= 0.75 as a descriptive working threshold, six of fifteen \texttt{crop} fields exceed the threshold while nine do not; on 49.66\% of garments all fifteen fields are simultaneously correct. The benchmark does not establish that exceeding this threshold is sufficient for unattended production use: the threshold is a reading aid, not a certification.

\section{What the \texttt{crop} score is made of}

Three measurements on the \texttt{crop} track's design, run on its frozen gold with the suite's own scorer and its 10,000-sample paired bootstrap clustered at the source image. None changes a number in Section 5; each changes how that number should be read. The intervals come from our own re-run of the bootstrap and can differ from the published ones in the fourth decimal.

\subsection{The ranking depends on the aggregation}

Six systems have committed \texttt{crop} predictions: MODA, the FashionSigLIP encoder ablation, the two zero-shot baselines, and two earlier systems of ours that fine-tune the generative Florence-2 model [16] to emit attributes as text (\texttt{parent}, and \texttt{corrective}, trained further to complete sparsely labelled fields). They give fifteen pairs. Attribute micro-F1 and field-macro F1 disagree on the leader in two of them, and in both the choice of aggregation decides whether there is a significant difference at all.

\begin{center}\small
\begin{tabular}{L{0.254\linewidth}L{0.186\linewidth}L{0.186\linewidth}}
\toprule
Pair (A minus B) & micro-F1 & field-macro F1 \\
\midrule
MODA minus FashionSigLIP encoder ablation & \textbf{+0.0056} [+0.0014, +0.0098] & $-$0.0019 [$-$0.0077, +0.0040] \\
Qwen3-VL-8B minus FashionSigLIP zero-shot & $-$0.0013 [$-$0.0080, +0.0056] & \textbf{+0.0207} [+0.0110, +0.0305] \\
\bottomrule
\end{tabular}
\end{center}

Micro-F1 pools decisions across fields, so fields with many gold values and predictions dominate it; field-macro gives every field equal weight. The other thirteen pairs agree under both. The \texttt{crop} headline in Section 5 is accurate as labelled, because micro-F1 was fixed before the comparison was run, but on field-macro MODA and the encoder ablation are statistically indistinguishable.

\subsection{Two thirds of the judged cells carry no annotation}

The track judges every garment on all fifteen fields, and scores a prediction on a field with no gold value as a false positive. Only \textbf{33.0\%} of those judged cells carry an annotation (23,234 of 70,320). Coverage runs from 100\% for category to 3\% for material. Re-scoring each garment only on the fields its annotation covers gives an upper bound, since it also forgives genuine hallucinations:

\begin{center}\small
\begin{tabular}{L{0.186\linewidth}rrr}
\toprule
System & micro-F1, as scored & annotated fields only & difference \\
\midrule
MODA (\texttt{moda-ner-v-crop}) & 0.6300 & 0.6764 & +0.046 \\
FashionSigLIP encoder ablation & 0.6245 & 0.6737 & +0.049 \\
parent & 0.5832 & 0.6484 & +0.065 \\
corrective & 0.4111 & 0.6250 & +0.214 \\
FashionSigLIP zero-shot [5] & 0.1817 & 0.2987 & +0.117 \\
Qwen3-VL-8B zero-shot [7] & 0.1805 & 0.2866 & +0.106 \\
\bottomrule
\end{tabular}
\end{center}

The penalty is not a constant offset: it is 0.046 for MODA and 0.214 for \texttt{corrective}, a system that emits more fields. Rankings within each column are preserved, but gaps compress. On annotated fields only, MODA minus the encoder ablation is +0.0027 [$-$0.0015, +0.0071] on micro-F1 and $-$0.0049 [$-$0.0107, +0.0011] on field-macro: neither excludes zero.

\subsection{Most of the score rests on absence decisions}

The released \texttt{crop} model emits a field only when a per-field applicability head, with thresholds calibrated on development data, judges it present. We ran the model once over all 4,688 crops and decoded three ways, changing only that threshold. Our re-run reproduces the published predictions on 4,598 of 4,688 rows (micro-F1 0.6295 against 0.6300); the remaining rows sit at threshold edges under different hardware numerics.

\begin{center}\small
\begin{tabular}{L{0.236\linewidth}rr}
\toprule
Decoding & micro-F1, as scored & annotated fields only \\
\midrule
shipped thresholds & 0.6295 & 0.6758 \\
applicability 0.5, uncalibrated & 0.6262 & 0.6849 \\
applicability off, every field emitted & \textbf{0.2696} & 0.7022 \\
\bottomrule
\end{tabular}
\end{center}

Switching the applicability decision off costs 0.36 of the 0.63 headline. The annotated-only column cannot say whether those decisions are right: every cell in it has a gold value, so there is no absent case to test, and suppressing a field can only cost recall. That column favours emitting everything by construction.

What the table does show is the boundary of the benchmark. Most of the \texttt{crop} headline is earned by deciding which fields to leave out; Fashionpedia's attribute labels [1] record no explicit absence and leave two thirds of judged cells unannotated. The track therefore cannot distinguish a correct absence decision from a learned habit of the source annotators. That is the measured case for independent annotation that records absence explicitly, which the \texttt{fullbody} track's explicit not-applicable state (Section 3.1) provides and the \texttt{crop} track does not.

\section{What does not improve these baselines}

Six interventions were run against the \texttt{fullbody} route without new labels. All six lost, and we report them because a negative result nobody can find gets re-run.

All were scored on the same 5,000-row frozen test split. They differ in what they were compared \emph{against}, and that distinction governs how each row reads. Interventions 1, 2, 3 and 6 modify a probe fitted on 1,000 calibration rows and are compared with the frozen-trunk probe fitted on the same 1,000 rows (tier-1 0.4534). Interventions 4 and 5 are complete systems and are compared with the shipped route, whose heads were fitted on 7,000 training rows (tier-1 0.6917). The two baselines are far apart because of training data, not architecture, so a delta is only ever meaningful against the baseline in its own row.

\begin{center}\footnotesize
\begin{tabular}{lL{0.422\linewidth}L{0.146\linewidth}rrL{0.097\linewidth}}
\toprule
\# & Intervention & Compared against & tier-1 & Delta & Paired 95\% CI \\
\midrule
1 & \texttt{crop} trunk transferred to full-body probes & frozen trunk, 0.4534 & 0.4499 & $-$0.0035 & [$-$0.0049, $-$0.0021] \\
2a & Gradient-boosted head, identical features & logistic regression, 0.4534 & 0.3754 & $-$0.0780 & [$-$0.0842, $-$0.0708] \\
2b & MLP head, one hidden layer of 256 & logistic regression, 0.4534 & 0.3540 & $-$0.0993 & [$-$0.1051, $-$0.0924] \\
3 & Late interaction: learned query over 196 patch tokens per field & pooled LR, 0.4534 & 0.4203 & $-$0.0331 & [$-$0.0370, $-$0.0288] \\
4 & Hybrid with FashionCLIP 2.0 & shipped route, 0.6917 & 0.6243 & $-$0.0673 & [$-$0.0751, $-$0.0597] \\
5 & Joint training of one trunk on two corpora & shipped route, 0.6917 & 0.5562 & $-$0.1355 & [$-$0.1444, $-$0.1248] \\
6 & Concatenate two trunks' features & frozen trunk, 0.4534 & 0.4629 & \textbf{+0.0095} & n/a \\
6$'$ & \emph{Capacity control:} random projection of the frozen features to the same width & frozen trunk, 0.4534 & 0.4629 & \textbf{+0.0096} & n/a \\
\bottomrule
\end{tabular}
\end{center}

Two rows carry most of the information. Row 5 is the only intervention that added new supervision rather than rearranging frozen artifacts; its training loss fell cleanly from 34.0 to 11.4, and it still lost 0.1355 tier-1 F1, degrading on precisely the conditional fields the shipped route handles with a calibrated applicability threshold. Optimisation was not the problem.

Row 6 was a positive result, and row 6$'$ is why it is not in the abstract. Concatenating a second encoder's features raised tier-1 from 0.453358 to 0.462857. Replacing that second encoder with a random projection of the \emph{original} features to the same width scored 0.462919, marginally higher. The gain was the added width, not complementary information, and we retracted it. Row 4 was abandoned rather than tuned: the field-level table showed FashionCLIP ahead on one field of eighteen, so any fitted mixing weight would collapse toward the pure MODA route.

\section{Observation: attribute vocabulary as a failure mode}

On 1,110 held-out images of the Stanford Clothing Attributes set, labelled by six MTurk workers per item and untouched during our development, we scored two of our own routes on the same five fields. A text-prototype route over the MODA encoder scores 0.716869 five-field macro set-F1; our supervised composite scores 0.617564. Paired delta 0.099305, 95\% interval [+0.085678, +0.112971]. (Version 1 reported 0.576394 for the composite; the correction is described at the end of this report.)

One field accounts for more than the whole deficit. The composite is the \emph{stronger} system on pattern (0.790991 against 0.720420), on sleeve length (0.940131 against 0.910196) and on collar presence (0.821147 against 0.583802), and the weaker on colour (0.483483 against 0.556757). It collapses on neckline: \textbf{0.052067 against 0.813170}.

The mechanism is less a lossy mapping than a vocabulary with the wrong partition. The composite's neckline head is trained on Shopping100k and returns \texttt{square} on 1,242 of the 1,856 images, while this evaluation partitions neckline into v-neck and round only. The head's vocabulary does include round and a low v-neck value, but its dominant answer has no slot here. The head is not so much wrong as answering a different question. The aggregate score obscures the source of the failure. Figure 3 shows four held-out examples with the gold label and both systems' predictions.

\begin{figure}[!tbp]
\centering
\includegraphics[width=\linewidth]{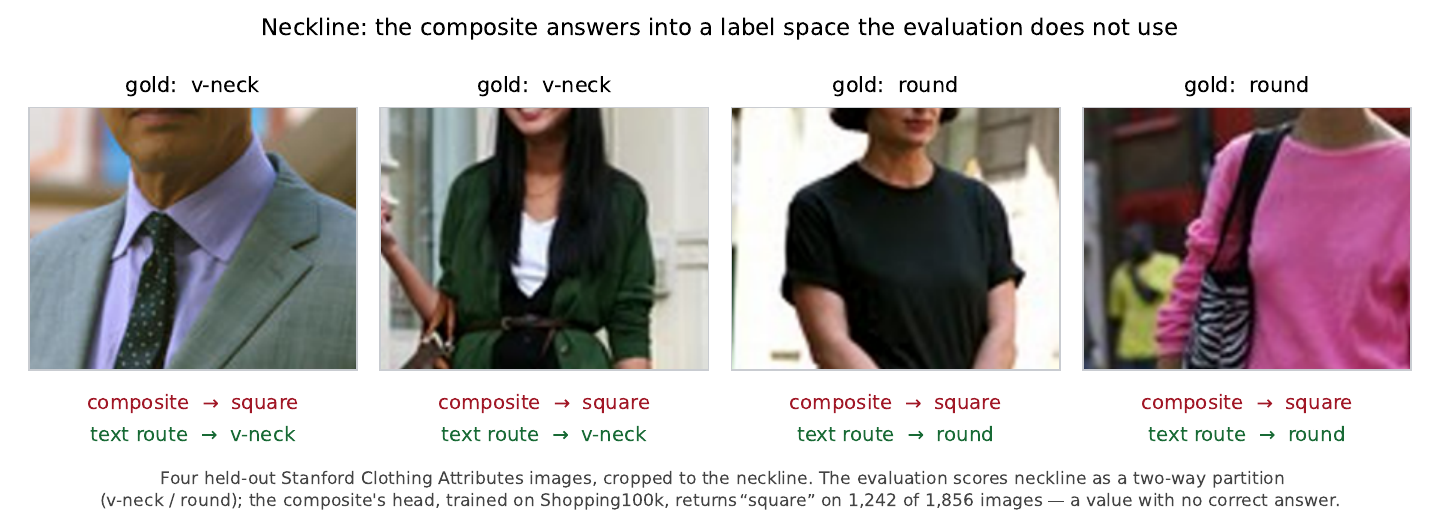}
\caption{Four held-out images, with each system's actual committed prediction. The composite answers in the Shopping100k vocabulary, and its dominant prediction, \texttt{square}, returned on 1,242 of the 1,856 images, has no slot in this evaluation's two-way partition and so no correct answer to be scored against. 315 held-out images show exactly this pattern: gold is a scored neckline value, the composite says \texttt{square}, and the text-prototype route is right.}
\end{figure}

Substituting the text route's neckline score into the composite's five-field macro, arithmetic on the per-field numbers above rather than a new run, gives 0.7698, which would place the composite 0.053 \emph{above} the route it currently loses to. One field's vocabulary mismatch inverts the ranking of two systems.

Neither route here is part of the released suite; both are internal packages over the public encoder, and this evaluation set is not one of the four tracks. The Stanford Clothing Attributes images are reproduced under the depositor's access condition that content not be used to identify individuals: Figure 3 is cropped to the garment neckline, and no face is shown. We record the observation and its magnitude. Methods for reconciling mismatched label spaces exist [12], and cross-domain attribute recognition is long established [11]; what we are not aware of is prior work reporting vocabulary mismatch as a scored result on a frozen fashion-attribute benchmark, with the ranking inversion made explicit. We have not yet measured whether the effect holds for other attributes or corpus pairs.

\section{Limitations}

\begin{center}\small
\begin{tabular}{L{0.407\linewidth}L{0.481\linewidth}}
\toprule
Limitation & What it means for the numbers above \\
\midrule
Labels are dataset-native, not independent human gold & Every figure derives from annotations each source corpus produced for its own purposes, not from annotators checking these claims. 1,993 image groups have been selected for independent annotation; that work has not started. Where a language model has adjudicated labels, we say so and treat it as diagnostic. \\
The comparator set is small & FashionCLIP 2.0, FashionSigLIP, Qwen3-VL-8B [7], and Gemini 3.5 Flash (\texttt{google/gemini-3.5-flash}), which was run against a continuation rule fixed before the run, failed it on the first 100 rows, and was stopped rather than completed. \\
The \texttt{crop} track's labels are sparse and record no absence & 33.0\% of judged cells are annotated, and most of the headline rests on absence decisions the labels cannot adjudicate (Section 6). Margins on this track are small and aggregation-dependent. \\
The \texttt{crop} track supplies localization & It isolates attribute prediction given an oracle garment box. A system that must locate the garment first will score lower than these numbers suggest. \\
The DeepFashion-MultiModal published protocol is not exactly reproduced & The split IDs used by [8] are not released. We construct a product-group-disjoint split and state throughout that our \texttt{fullbody} numbers are not comparable to the published table. \\
The \texttt{text} track's labels are silver & Opened during development, and 4 of 13 entity types are weak, unmeasurable or unscored. Its weights are not distributed. \\
Section 8 sits outside the suite & Its evaluation set is not one of the four tracks and neither route it compares is released. Its labels are, however, the strongest in this report, six independent MTurk annotators per item, versus dataset-native labels everywhere else. It is a single observation on one attribute and one corpus pair, and we do not present it as a study. Its labels had been opened before the version 2 correction, which changes preprocessing only. \\
Two tracks depend on research-only corpora & Their terms extend to derived data, so the weights trained on them are non-commercial. We state the licence rather than seek a permissive one. \\
\bottomrule
\end{tabular}
\end{center}

\section{Availability}

Code, scorers, split builders, prediction files and hashes: \texttt{github.com/hopit-ai/Moda\_ner}. Tables and protocol: \texttt{hopit-ai.github.io/Moda\_ner}.

Released weights, all on Hugging Face: \texttt{moda-ner-v-crop} (MIT), \texttt{moda-ner-v-catalog} and \texttt{moda-ner-v-fullbody} (CC BY-NC 4.0). All three are heads and adapters over \texttt{HopitAI/moda-fashion-distilled}, our own MIT-licensed encoder, which uses a \texttt{ViT-B-16-SigLIP} architecture [6]; no third-party encoder is loaded at inference.

\textbf{Not released, and named here so the boundary is unambiguous:} the text model behind the 0.8723 figure, the hosted routing runtime, and the internal composite routes compared in Section 8.

\textbf{What can be re-scored from shipped files.} Prediction files with committed hashes ship for every image-track system in the Section 5 table, MODA and comparator alike, and for all six \texttt{crop} systems in Section 6, including the two zero-shot baselines. The three Section 6.3 re-decodings ship with their hashes, and the Section 6 analysis scripts ship in \texttt{suite/crop/analysis/}, so every image-track number in Sections 5 and 6 can be recomputed from the repository. The following are reported from our run records and cannot yet be re-scored independently: the \texttt{text} prediction file, the Section 7 intervention runs, and the Section 8 route predictions.

\section{Changes in version 2}

\textbf{Correction.} The released inference code for the \texttt{crop} and \texttt{fullbody} models (\texttt{models/inference.py} in the repository) built their image transform for the wrong encoder configuration: OpenAI-CLIP normalisation with a centre crop, instead of the mean and standard deviation of 0.5 with squash resizing that the checkpoints were trained with. The Section 5 numbers were produced by a correctly configured path and are unchanged. On 64 held-out images each, the fixed code reproduces the published predictions on 64 \texttt{crop} and 63 \texttt{fullbody} images, against 16 and 0 before; a regression test now guards the transform. One figure in version 1 was affected. In Section 8 (Section 7 in version 1) the supervised composite takes its collar field from the \texttt{crop} model through the same construction. Re-run with the correct preprocessing, with the new predictions committed by hash before scoring and the scorer unchanged except for that hash, its collar F1 moves from 0.615298 to 0.821147 and its five-field macro from 0.576394 to 0.617564; no other field changes. The paired delta becomes 0.099305 (version 1: 0.140475), and the substitution arithmetic 0.7698 (version 1: 0.7286). The finding stands and the ranking inversion widens. The labels had already been opened, so this is a defect correction with the same weights and thresholds, not a fresh held-out result.

\textbf{Additions.} Section 6, measuring the \texttt{crop} track's dependence on aggregation, annotation sparsity and applicability decisions; the field-macro row in the Section 5 table.

\textbf{Wording.} Section 8 no longer calls the two neckline vocabularies non-intersecting: the composite's vocabulary contains round and a low v-neck value, and it is its dominant answer that has no slot. The Availability section now states exactly which prediction files ship, where version 1 said every number was reproducible from shipped files; the zero-shot \texttt{crop} predictions and the Section 6 files were added to the repository with this version. The limitations now state that independent annotation has not started, and drop a statement that comparator expansion was in progress. Cross-references and a subsection number are corrected.

\end{document}